\pdfoutput=1

\documentclass[11pt]{article}

\usepackage[preprint]{acl}

\usepackage{times}
\usepackage{latexsym}

\usepackage[T1]{fontenc}

\usepackage[utf8]{inputenc}

\usepackage{microtype}

\usepackage{inconsolata}

\usepackage{enumitem}
\usepackage{amsthm}
\theoremstyle{definition}
\newtheorem{definition}{Definition}

\usepackage{graphicx}

\title{Biomedical Question Answering via Multi-Level Summarization on a Local Knowledge Graph}

\author{Lingxiao Guan \\
  University of Michgian \\
  \texttt{lxguan@umich.edu}\\\And
  Yuanhao Huang \\
  University of Michigan \\
  \texttt{hyhao@umich.edu}\\\And
  Jie Liu \\
  University of Michigan \\
  \texttt{drjieliu@umich.edu}}

\begin{document}
\maketitle
\begin{abstract}
In Question Answering (QA), Retrieval Augmented Generation (RAG) has revolutionized performance in various domains. However, how to effectively capture multi-document relationships, particularly critical for biomedical tasks, remains an open question. In this work, we propose a novel method that utilizes propositional claims to construct a local knowledge graph from retrieved documents. Summaries are then derived via layerwise summarization from the knowledge graph to contextualize a small language model to perform QA. We achieved comparable or superior performance with our method over RAG baselines on several biomedical QA benchmarks. We also evaluated each individual step of our methodology over a targeted set of metrics, demonstrating its effectiveness.
\end{abstract}

\section{Introduction}

Multi-Document Question Answering (MDQA) via Large Language Models (LLMs) has improved the experience and effectiveness of users by generating relevant, informed responses. A key approach in MDQA is Retrieval-Augmented Generation (RAG) \cite{DBLP:conf/nips/LewisPPPKGKLYR020}, in which LLMs are augmented with documents retrieved from established corpora providing up-to-date information. RAG reduces hallucinations due to the retrieved contexts grounding the answer, and providing access to domain-specific knowledge in knowledge bases. These benefits are even more important in the domain of BioMedical Question Answering (QA), as it has much higher requirements in terms of what data is used and the outputs of models. Many questions require the integration of information from multiple documents, as healthcare professionals may use model outputs to inform decisions such as medical diagnoses of rare and complex conditions.

However, extracting and leveraging multi-document relationships remains an underexplored challenge. Directly placing retrieved documents into an LLM’s context window fails to capture inter-document connections, thereby limiting the model's ability to aggregate evidence and leading to inadequate reasoning. This issue is particularly pronounced in Biomedical QA, where accurate answers often require synthesizing multiple medical concepts across diverse documents. While prior work has introduced RAG-based techniques to mitigate the limitations of finite context windows and noisy, inconsistent information in LLMs \cite{DBLP:journals/corr/abs-2312-10997}, these methods often overlook the deeper relationships between retrieved documents. A recent approach aimed at addressing this problem in unstructured knowledge base involves hierarchical summarization of semantically related chunks \cite{DBLP:conf/iclr/SarthiATKGM24}. However, because \emph{documents can share topics while differing in semantic focus}, this method risks missing critical connections and producing non-diverse summaries. For example, a document discussing breast cancer outcomes and another on its genetic causes may both be relevant to a query about breast cancer, but may not be identified as related due to their differing semantic focal points. Knowledge Graphs (KG) have been proposed as an alternative for their representation of explicit relationships between concepts. However, existing KG-based methods exhibit key limitations. Some approaches require access to an entire offline knowledge corpus \cite{DBLP:journals/corr/abs-2404-16130}, limiting their applicability to dynamically retrieved documents. Others construct graphs dynamically but suffer from explicit information loss during graph traversal for retrieval \cite{DBLP:conf/aaai/0160LRSZD24}. Therefore, there is a need for a method that effectively extracts and utilizes relevant multi-document relationships from dynamically updated knowledge bases, enabling more comprehensive reasoning in Biomedical QA and beyond. 

To remedy this, we propose \textbf{utilizing the construction of a knowledge graph to underlay layerwise document summarization} as an alternative. We utilize propositional claims to represent information and facilitate handling conflicting and noisy claims extracted from unstructured documents we retrieve. The knowledge graph structure constructed from these propositional claims captures relationships beyond semantic similarity. Finally, our approach performs layerwise graph summarization around several key claims of interest to comprehensively capture multi-document relations and fit them into a limited context window.

This method utilizes the properties of decontextualized claims in the knowledge graph structure and layerwise topological summarization to capture explicit and implicit relationships between entities in the documents, thus having a more comprehensive context to provide to LLMs. We evaluate each part of our methodology, and compare our method to traditional RAG retrieval baselines on several biomedical QA datasets, achieving comparable or superior performance over all baselines.

Our approach makes three main contributions.
\begin{itemize}
    \item We introduce a novel method of structuring the information from retrieved documents as propositional claims in local knowledge graphs.
    \item We introduce a technique utilizing layerwise topological graph summaries of key claims in this local knowledge graph as context for LLM QA tasks.
    \item We evaluate our approach on a comprehensive set of benchmarks, testing both the properties of the intermediate results of the method and the final accuracy on several datasets.
\end{itemize}

\section{Related Works}

\subsection{Retrieval Augmented Generation}
Information Retrieval methods have long been used for general question answering tasks, including biomedical QA \cite{10.1145/3490238}. RAG extends these methods for use with LLMs, allowing for the integration of large external corpora into pre-trained Language Models. The naive RAG approach was first introduced in Lewis et al. \cite{DBLP:conf/nips/LewisPPPKGKLYR020}, and has since been followed by many follow-up refinements \cite{DBLP:journals/corr/abs-2312-10997}. A number of works have been conducted on the application of RAG in biomedical QA, such as MedRAG which retrieves documents from a variety of corpora \cite{DBLP:conf/acl/Xiong0LZ24}, BioMedRAG which trains the retriever for improved retrieval of medical documents \cite{li2024biomedrag}, and Self-BioRAG which uses on-demand retrieval and reflection tokens to select the best evidence \cite{pmlr-v225-yu23b}, among many others \cite{Liu2024ASO, Zhou2023ASO}, which tend to take the strategies used in general domain RAG and adapt then to the biomedical domain. While naive RAG provides benefits for QA tasks, a number of follow up works extend it. 

\subsection{Summarization}
Summarization of the input contexts is one method by which the retrieved documents can be further processed to better suit downstream tasks. Summarization, by its nature, is capable of condensing input documents into a format that retains relevant information while using less input tokens. RAPTOR \cite{DBLP:conf/iclr/SarthiATKGM24} attempts to use hierarchical summarization of input documents to capture both locally relevant information and distant interdependencies through multiple summarization levels. However, its reliance on semantic similarity means that it may miss explicit, non-semantic connections. Long-context summarization methods like MemTree \cite{DBLP:journals/corr/abs-2410-14052} follow a similar vein of using embedding similarity to group contextual information, and thus suffer from the same problem of missing explicit connections. 

\subsection{RAG with Knowledge Graphs}
More recently, there has been a line of work attempting to perform community-based summarization on generated knowledge graphs. They partition the knowledge graph into modular parts, either via communities as with Graph Rag \cite{DBLP:journals/corr/abs-2404-16130}, or into hierarchical tags as in MedGraphRAG \cite{DBLP:journals/corr/abs-2408-04187}. While these methods are able to capture more multi-document relationships, they perform their method on the entire offline retrieval corpus rather than dynamically retrieved online input documents. This requires a high upfront cost and a different level of granularity compared to our method, while also requiring additional effort to update their graph summaries with new information. 

Alternatively, retrieved documents can be turned into a graph structure for additional processing. Several works have opted for this method, with many using semantic similarity of text chunks or a combination of semantic similarity and structural information \cite{DBLP:conf/aaai/0160LRSZD24, DBLP:journals/corr/abs-2408-02907} to construct the graph. These methods that use semantic similarity are unable to capture explicit connections, and even with explicit connections formed by structural relationships the retrieval uses agents that can miss information outside of the explicitly returned paths. Our method utilizes the explicit connections formed from knowledge graph RDF formats and does layerwise summarization to capture these connections and relevant pieces of information.

\begin{figure*}[h]
    \centering
    \includegraphics[width=1\textwidth]{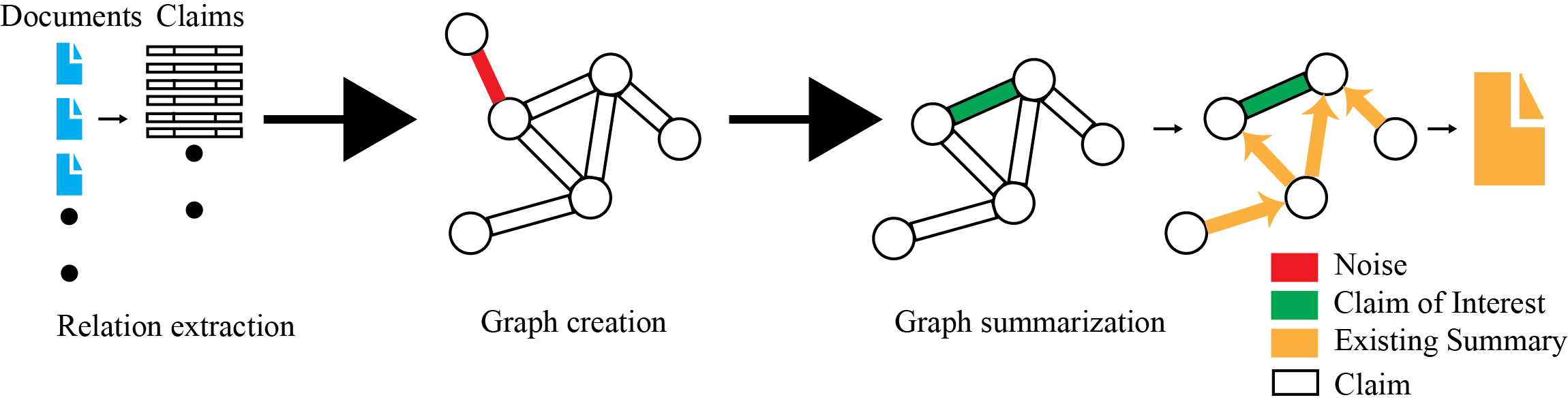}
    \caption{Overview of the proposed layerwise summarization method. \textbf{Relation extraction:} Load in documents with a retriever, break documents into claims, break claims into triples. \textbf{Graph creation:} Build local graph with triples and denoise. \textbf{Graph summarization:} Summarize the graph layerwise with the top re-ranked claims as the roots. The final summaries are provided to a model as context for downstream QA tasks.}
    \label{fig:mesh1}
\end{figure*}

\section{Methods}
\textbf{Approach Overview}: Our approach handles the problem of processing and connecting information from multiple retrieved documents to solve biomedical questions. At its core, our methodology takes in a biomedical question, a set of retrieved documents, and possible multiple choice answers before using a language model to process the documents and determine the correct answer. More formally, given an input biomedical question \textit{q}, a set of answer options \textit{A}, and a corpus of dynamically updated unstructured documents \textit{D}, a language model \textit{L} is used to select the correct answer \textit{a $\in$ A}. The output should satisfy three requirements:
\begin{enumerate}[noitemsep,leftmargin=*]
    \item Comprehensively identify and connect multi-document relations.
    \item Efficiently use the limited context window of \textit{L}.
    \item Reduce noise while preserving relevant information.
\end{enumerate}
Our method seeks to improve the extraction and presentation of relevant information and multi-document relations from unstructured documents by the addition of layerwise graph summarization, and the process can be seen in Figure \ref{fig:mesh1}. It proceeds by first extracting decontextualized claims from each \textit{d} $\in$ \textit{D} (Section \ref{sec:relextract}), using the entities in these claims to build a graph (Section \ref{sec:graphcreate}), before summarizing the content in the graph topologically into several key claims that are provided to \textit{L} to solve the question (Section \ref{sec:graphsumm}). 

\subsection{Relation Extraction}\label{sec:relextract}
The relation extraction step transforms retrieved unstructured documents into propositional claims and associated RDF Triples. This allows us to transform complex technical documents into atomic pieces of information that can be reliably connected and analyzed in later steps. It is subdivided into three sections, including retrieving the unstructured documents from several knowledge bases, extracting and decontextualizing propositional claims from these unstructured documents, before extracting RDF Triples from each of these claims to facilitate graph construction.
\newline

\noindent
\textbf{Retrieval}\label{sec:retrieval}: To accurately answer biomedical questions, our approach gathers relevant information from several knowledge bases. For a given input question \textit{q}, the input question is first preprocessed into a better suited retrieval query to retrieve relevant documents $d \in D$. As mentioned in previous work \cite{DBLP:journals/corr/abs-2305-14283}, question rewriting helps to better match the input question to the model's understanding, improving the clarity and effectiveness of the questions. 

Our method follows with HyDE candidate answer generation \cite{gao-etal-2023-precise}. This approach generates a candidate answer using the model \textit{L} given the input question \textit{q} and associated answer options. The candidate answer includes domain-specific terminology and context from the model's internal parameters, which the retriever uses alongside the input question and answer options to capture documents that share similar concepts and improve the retrieval results.

The final query with the rewritten question, answer options, and candidate answer is used to retrieve text chunks $d \in D$ from four knowledge bases: Statpearls documents, Simple Wikipedia, Medical Textbooks, and PubMed Abstracts/Fulltext articles. Further details on the retrieval corpora and the retrieval process can be found in Section \ref{sec:RAGDatasets}.
\\

\noindent
\textbf{Claim Extraction}\label{sec:claimExtract}: To connect information across documents, documents are broken down into concise and independent pieces. From the retrieved text chunks $d \in D$, we extract propositional claims $C = \{c_1, c_2, ..., c_n\}$ with the model $L$. These propositional claims must be
\begin{itemize}[noitemsep,leftmargin=*]
    \item Atomic: includes only a single statement that cannot be broken down, and
    \item Decontextualized: fully understandable on its own with no unresolved entity references.
\end{itemize}
A previous work \cite{DBLP:conf/emnlp/Chen0C0MZ0024} has found that using this chunking modality improves retriever performance. The decontextualization process is especially important for our task, because in later reranking and summarization our approach needs to not only understand the meaning of each claim in isolation, but also its importance to the input question without additional context.\\

\noindent
\textbf{Triple Extraction}\label{sec:tripleExtract}: Once each claim $c \in C$ is extracted, we need to prepare them for addition to the local graph $G$ that will be constructed. We assume that the decontextualization and claim extraction process has given us atomic propositional claims, with each one only having one key relation. Our method involves extracting a single RDF triple $(subj,\ pred,\ obj)$ from each claim $c$. This triple format captures the relationship $pred$ between the two entities $subj$ and $obj$.  

\subsection{Graph Creation}\label{sec:graphcreate}
The graph creation step processes the RDF Triples and claims from the relation extraction step into a local graph structure that captures the relationships between pieces of information. This is crucial for identifying multi-document interactions that are not apparent from individual claims. It can also be subdivided into several key steps, including deduplication of RDF triple entities, constructing the local graph, and denoising irrelevant information from the KG.
\newline

\noindent
\textbf{Deduplication}\label{sec:dedup}: While our claim extraction phase (Section \ref{sec:claimExtract}) resolves coreferences to the same entities, the entities in each RDF triple can still have multiple possible representations. This means that minor character differences can result in different entity nodes for the RDF triples, breaking apart connections.

Deduplication of entities in the RDF triples is performed to ensure that all references to the same concept point towards the same node in the graph. Our approach uses the cosine similarity of each entity's embeddings rather than character-based Levenshtein distance because medical entities that have only minor character differences can have entirely different meanings. Specifically, embeddings are placed into the same cluster with Unweighted Average Linkage Clustering (UPGMA) \cite{sokal1958statistical}, using the average embedding similarity of each entity in the cluster. Using a similarity threshold of 0.8, we map all entities within a cluster to an arbitrary label in the cluster. This restores the connections between claims that discuss similar concepts.  
\\

\noindent
\textbf{Graph Construction}\label{sec:graphConstruct}: After deduplication, the processed RDF triples and claims are used to construct the graph $G$. Each node in the graph is an entity from the RDF triples $(subj, \ pred, \ obj)$, one of the $subj$ or $obj$ entities. Each edge in the graph represents the relationship between the two entities and includes the representative claim $c$ and relevancy score $s$. Each of the relevancy scores are calculated using a reranker $R$ according to how much the edge claim addresses the input question \textit{q}. All of the edges are treated as undirected in further processing, and allow for multiple edges between two entities. 
\\

\noindent
\textbf{Denoising}\label{sec:denoise}: While nodes that represent the same concept have been grouped together, there remains the possibility that noise has been added to $G$. Noise is a common problem in knowledge graph creation, and it can mislead the LLM when answering the input question or crowd out relevant information from the model's context window.

We consider a claim $c$ to be noise if it is both
\begin{itemize}[noitemsep,leftmargin=*]
    \item Irrelevant to the input question directly, or
    \item Not useful in connecting claims that are important to answering the question.
\end{itemize}
While using a reranker can determine claim relevance, it is insufficient to determine whether the claim helps in connecting important information. Following \cite{DBLP:journals/corr/abs-2409-00861}, our approach evaluates claim importance using its connections in $G$. Specifically, our method provides each claim's 1-hop connected neighbors as context for a model \textit{L} to determine the claim's relevance to the input question. 

\subsection{Graph Summarization} \label{sec:graphsumm}
The final graph summarization stage of our methodology condenses the content in $G$ into several key claims of interest to capture the most relevant information for answering the input question. This stage involves first obtaining claims of interest as entry points into the graph, performing layerwise summarization to capture relevant connections between the claims, before generating focused summaries around the claims of interest to obtain a context that is provided to the model for answering the input biomedical question.
\newline

\noindent
\textbf{Obtaining Claims of Interest}\label{sec:claimsOfInterest}: Due to the number of documents under consideration, we need to focus on only the most relevant information. Our method selects several key claims of interest \textit{K} from $G$, which will provide a diverse set of entry points into the graph. Our approach starts with the top 10 ranked claims in the graph, because individually they were determined by the reranker to be the most relevant for answering the question.

We want to determine each claim of interests' potential to produce meaningful summaries for our later layerwise summarization phase. Since claims closer to the claims of interest will be given more weight in the final summaries, each claim of interest's 1-hop neighboring claims are examined. These neighboring claims are used as context to generate \textit{test summaries}, and the ranks of these test summaries are used to again rerank the claims of interest.

To ensure that there is sufficient diversity in perspectives, these reranked claims are filtered. As adjacent claims should produce similar summaries, we merge all claims that are a 1-hop neighbor of a higher ranked claim into that higher ranked claim, removing it from $K$. This returns a shorter, more focused list of claims as our final set for \textit{K}, improving the efficiency of the subsequent procedures while retaining coverage of relevant information.

\begin{figure}[h]
    \centering
    \includegraphics[width=0.45\textwidth]{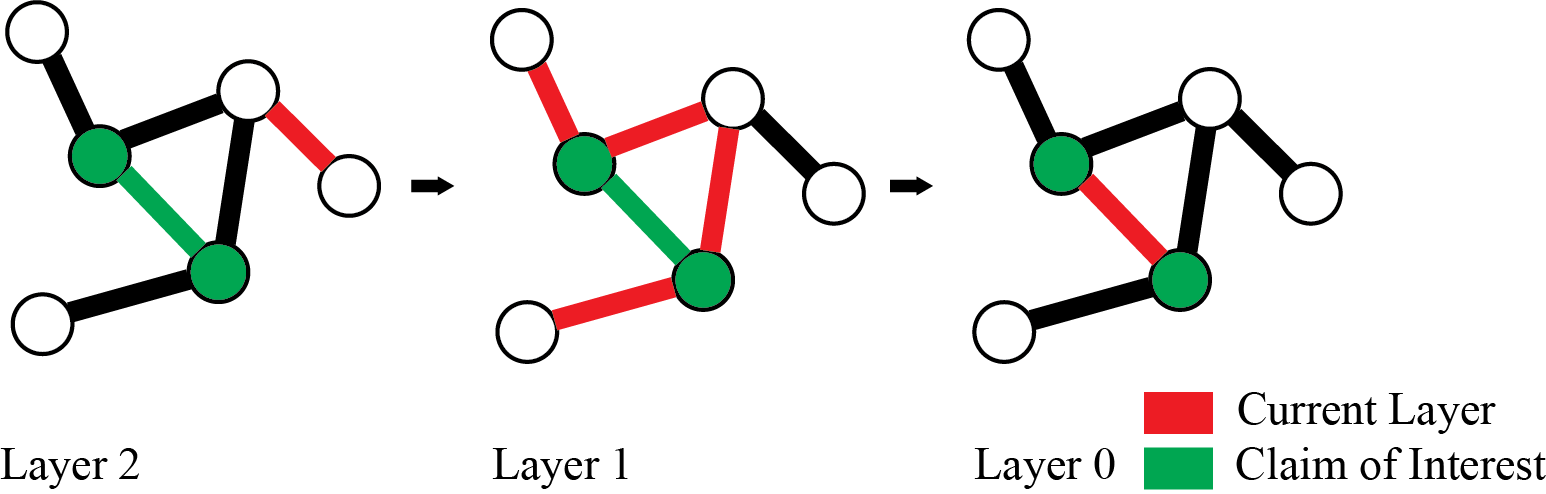}
    \caption{Layerwise summarization method overview. For a given claim of interest, the graph is organized into layers based on the distance of each connected claim from the claim of interest. The summarization process begins from the furthest layer, moving inwards. For each layer claims are summarized using the previously generated summaries of their connected claims in lower layers. This process ensures that path information and multi-document relationships are preserved while filtering out irrelevant information in the final summaries. }
    \label{fig:layerwise}
\end{figure}

\noindent
\textbf{Layerwise Summarization}\label{sec:layerwiseSum}: The process of layerwise summarization for each claim of interest involves organizing its connected component in $G$ into layers based on the distance each claim is from it.

\begin{definition}[Layer]
    Given a claim of interest $k$ in graph $G$, the $i$th layer consists of all claims that are exactly $i$-hop away from $k$ in $G$.
\end{definition}

The summarization process starts from the outermost layer and proceeds inwards. For each claim in the current layer, our method considers the summaries of connected claims one layer below. These summaries from connected claims are again summarized to create the current claim's own summary. Each claim is processed only once and only uses summaries from already processed claims, ensuring that there are no cycles. This occurs layer by layer until the claim of interest is reached. An example visualization of our method can be seen in Figure \ref{fig:layerwise}. 

This layerwise summarization is used because it has three key benefits. First, it is capable of capturing all the information in the local connected component, including both the direct content and path-based information. This is important for understanding multi-document relations between different medical concepts. Second, our layerwise processing of claims will inherently filter out irrelevant content, complementing our denoising step. Finally, this method places emphasis on claims closer in $G$ to the claims of interest, which naturally prioritizes more topically relevant information in the final summaries. \\

\noindent
\textbf{Summary Generation}\label{sec:summaryGeneration}: The final summarization for each claim of interest captures the information from its entire connected component in $G$, but is focused on the perspective around the specific claim. While these claims share common topics due to their high relevance to the input question, each final summary should differ in content due to them emphasizing their local relationships. The final output of this method is a concatenation of all of the summaries in the order of their relevance rankings, from the highest to lowest. This set of summaries is provided as contexts for an LLM to perform QA tasks.

\section{Experiment Settings}

\subsection{Datasets}
\subsubsection{RAG Datasets}
\label{sec:RAGDatasets}
The retrieval corpora in this work are a combination of medical and general corpora, including 
\begin{itemize}[noitemsep,leftmargin=*]
    \item Simple Wikipedia \cite{wikidump}, 
    \item Medical textbooks from MedQA \cite{jin2020disease}, 
    \item PubMed abstracts and full text articles from GLKB \cite{huang2024building}, and 
    \item StatPearls articles\footnote{https://www.statpearls.com/}. 
\end{itemize}  

Simple Wikipedia provides general knowledge, medical textbooks provide foundational concepts, Statpearls documents provide detailed medical information, and PubMed abstracts/fulltext articles provide research findings. This combination is to improve the coverage of topics our method can retrieve relevant information for, inspired by MedRAG \cite{DBLP:conf/acl/Xiong0LZ24}. 
\subsubsection{Evaluation Datasets}
\label{sec:EvalDatasets}
Our evaluation datasets include the test sets of PubMedQA \cite{jin2019pubmedqa}, MedQA \cite{jin2020disease}, and the MMLU clinical topics datasets \cite{DBLP:conf/iclr/HendrycksBBZMSS21} (Anatomy, Clinical Knowledge, College Biology, Professional Medicine, College Medicine, and Medical Genetics). For validation and ablation tests, we utilize a combination of the validation sets of the individual MMLU datasets which we term MMLU Validation.
\subsection{Model Settings}
We use the Mistral-7B-Instruct-v0.1 model for both construction and summarization of the graph for all evaluations \cite{DBLP:journals/corr/abs-2310-06825}. For experiments that involved LLM-as-a-judge capabilities, we used Mixtral-8x7B-Instruct-v0.1 \cite{DBLP:journals/corr/abs-2401-04088}. For Reranking, we used bge-reranker-v2-gemma, and for embedding we used bge-large-en-v1.5 \cite{li2023making}. We use the en\_core\_sci\_scibert spacy model \cite{neumann-etal-2019-scispacy} due to its better performance on scientific tasks compared to general domain spacy models, and the neural entity recognition pipeline to extract entities.

\subsection{QA Baselines}
We compared the QA accuracy of our Layerwise method with four alternative measures.
\begin{itemize}[noitemsep,leftmargin=*]
    \item Baseline: The prompt only includes the input question and answer options, relying on the model's parametric knowledge to answer the questions.
    \item Rewrite: We use question rewriting to retrieve unstructured documents from the knowledge bases, before adding them with reranking to the model's context window until the context limit is reached.
    \item HyDE \cite{gao-etal-2023-precise}: We use the HyDE query generation method to use the question, answer options, and candidate answer to retrieve unstructured documents from the various knowledge bases. The retrieved documents are reranked and added to the model's context window up to the context limit.
    \item RAPTOR \cite{DBLP:conf/iclr/SarthiATKGM24}: We use the HyDE query generation method to retrieve documents. The RAPTOR process is used to produce a context for each question for QA.
\end{itemize}
We evaluated these methods on the benchmarks discussed in the Evaluation Datasets (Section \ref{sec:EvalDatasets}) Section.

\subsection{Component Level Analysis}
We evaluated the capabilities of each of the individual components in our methodology to determine their effectiveness and validate key assumptions over our MMLU Validation dataset. These included the modules of relation extraction, graph creation, and graph summarization as can be seen from Figure \ref{fig:mesh1}. 
\\

\noindent
\textbf{Relation Extraction}: The goal of the relation extraction phase is to turn the retrieved documents into decontextualized claims with associated RDF Triples. The desired properties of these claims and triples are that each claim is self-contained and the meaning of the source documents are retained. Thus, for relation extraction, we evaluated the methodology's ability on three key criteria, namely: \\
\begin{itemize}[noitemsep,leftmargin=*]
    \item Decontextualization: fraction of explicit entity references over all entity references extracted with SpaCy from each claim. 
    \item Preservation of semantic meaning: the semantic similarity between the embedding of the input document and the concatenated form of all of the extracted claims.
    \item Key claim extraction: the fraction of key claims extracted from the retrieved documents using a judge LLM that are retained in the output summaries.
\end{itemize}

To assess our method, we compare it with several alternatives. 
\begin{itemize}[noitemsep, leftmargin=*]
    \item Single stage (Our Method): Extracts the claims from the documents and decontextualizes them in a single prompt.
    \item Two stage: Performs the extraction and decontextualization separately, could potentially improve the performance of the decontextualization but has a drop in efficiency.
    \item Direct triples: Extracts RDF triples instead of claims, improves the efficiency of the overall pipeline due to skipping the claim extraction step.
    \item Pairs relations: Extracts the entities first before extracting the relations between entities, a more traditional KG creation method.
\end{itemize}

\noindent
\textbf{Graph Creation}: The goal of the graph creation phase is to have the RDF Triples from the relation extraction phase connect related claims. The communities in the graph should make sense upon consideration of their relevance to the input question. Thus, for graph creation, we tested the methodology's ability to \textit{have high quality graph communities centered around key claims.} 

We compared the summaries produced from subgraphs and semantic communities around the claims of interests we obtained from the graph summarization stage (Section \ref{sec:graphsumm}). 

\begin{itemize}[noitemsep, leftmargin=*]
    \item Subgraph communities: we consider all 1-hop connections around the entities in the claims of interests, using the claims on these connections to produce summaries for each claim of interest.
    \item Semantic communities: we retrieve all claims that have a similarity above the cosine similarity threshold of 0.8 with the claims of interests, and use these claims to produce summaries
\end{itemize}

The metric's score for an index with either method is calculated by obtaining the relevance score relative to the input question of the concatenation of all produced summaries of that index. As the actual relevance scores produced by rerankers are only useful to compare the two methods, we record which of the two methods had a higher score for each index.
\\

\noindent
\textbf{Graph Summarization}: The goal of graph summarization is to ensure that the summaries produced by the summarization method are useful for the input question. The requirements for these summaries are that the contents should be \textit{relevant}, \textit{have little hallucinations}, and \textit{have information from various sources}. 

Thus, for graph summarization, we further test three different metrics: 
\begin{itemize}[noitemsep,leftmargin=*]
    \item Faithfulness (hallucination rate): fraction of claims in the output summaries that are supported by the input documents.
    \item Answer relevance: fraction of claims relevant to the input question in the output summaries.
    \item Score diversity: fraction of input documents that are included in the final summaries.
\end{itemize} 

We compared our layerwise approach with the summaries produced from the graph communities formed from the 1-hop subgraphs around claims of interests and those produced with semantic communities around the claims of interests. These are the same summaries we used in the Graph Creation component analysis.

\section{Results}

\begin{table*}[h]
\begin{center}
\begin{tabular}{p{2.2cm}|p{1.15cm}|p{1.15cm}|p{1.15cm}|p{1.15cm}|p{1.15cm}|p{1.15cm}|p{1.15cm}|p{1.15cm}|p{1.15cm}}
    \hline
     Approach & MMLU-V\textsuperscript{*} & MMLU-A & MMLU-CB & MMLU-CM & MMLU-PM & MMLU-MG & MMLU-CK & PMQA & MedQA \\
     \hline
     Baseline & 0.55 & 0.46 & 0.57 & 0.46 & 0.51 & 0.6 & 0.54 & 0.50 & 0.44 \\
     Rewrite & 0.47 & 0.44 & 0.45 & 0.38 & 0.48 & 0.62 & 0.43 & 0.59 & 0.46 \\
     HyDE & 0.55 & 0.47 & 0.47 & 0.45 & 0.57 & 0.65 & 0.46 & 0.60 & 0.50 \\
     RAPTOR & 0.63 & 0.54 & 0.63 & 0.55 & 0.60 & 0.75 & 0.63 & \textbf{0.66} & 0.50 \\
     Layerwise & \textbf{0.63} & \textbf{0.55} & \textbf{0.65} & \textbf{0.56} & \textbf{0.62} & \textbf{0.78} & \textbf{0.65} & 0.56 & \textbf{0.54} \\
    \hline
     \multicolumn{10}{l}{\footnotesize \textsuperscript{*}MMLU prefixes denote: V-Validation, A-Anatomy, CB-College Biology, CM-College Medicine, PM-Professional Medicine, }\\
     \multicolumn{10}{l}{\footnotesize MG-Medical Genetics, CK-Clinical Knowledge}\\
    \end{tabular}
\end{center}
\caption{Comparison of accuracy scores across various BioMedical QA approaches. Results show the performance on MMLU Clinical Topics, PubMedQA, and MedQA benchmarks. Our Layerwise method shows consistent improvements over baseline methods, with comparable or superior performance across the non-validation datasets. The MMLU prefixes denote different subject areas, as noted under the table.  }
\label{table:QAAcc}
\end{table*}

\begin{table}[h]
\begin{center}
    \begin{tabular}{p{2.1cm}|p{1.5cm}|p{1.5cm}|p{1.1cm}}
    \hline
     Approach & Ref Score & Sem. Sim. & Claim Ret. \\
     \hline
     single\_stage & 0.941 & 0.901 & 1.0 \\
     two\_stage & 0.946 & 0.903 & 1.0 \\
     direct\_triples & 0.971 & 0.865 & 1.0 \\
     pairs\_relations & 0.994 & 0.815 & 1.0 \\
     \hline
\end{tabular}
\end{center}
\caption{Comparison of Relation Extraction methods across three metrics. Ref. Score measures decontextualization ability, Sem. Sim. measures preservation of original meaning, and Claim Ret. measures preservation of key information. Higher scores indicate improved performance on the individual metrics, ranging from 0-1.0. Results demonstrate the trade-off between entity-based and claim-based approaches, with our single stage method achieving a balanced performance while maintaining good computational efficiency.  }
\label{table:Rel}
\end{table}

\begin{table}[ht]
\begin{center}
    \begin{tabular}{c|c}
    \hline
     Approach & Summary Score Wins \\
     \hline
     Graph Communities & 59.35\% \\
     Semantic Communities & 40.65\% \\
     \hline
    \end{tabular}
\end{center}
\caption{Comparison of relevance scores between graph and semantic-based summarization. Results show the percentage of times each method produced summaries with a higher relevance score, and demonstrate the graph summary's superior ability to capture relevant information from the input documents. }
\label{table:GraphCrea}
\end{table}

\begin{table}[!ht]
\begin{center}
    \begin{tabular}{p{1.5cm}|p{1.7cm}|p{1.5cm}|p{1.5cm}}
    \hline
     Approach & Faithfulness & Relevancy & Source Diversity\\
     \hline
     Layerwise & 0.9569 & 0.8414 & 0.9647 \\
     Semantic & 0.9706 & 0.8604 & 0.9170 \\
     Subgraph & 0.9453 & 0.7938 & 0.9356 \\
     \hline
    \end{tabular}
\end{center}
\caption{Evaluation of three summarization approaches across faithfulness (hallucinations), relevancy (relevance to input question), and source diversity (multi-document relations) metrics. Scores range from 0-1.0. Results demonstrate the Layerwise summarization method's ability to maintain a high faithfulness and relevancy while achieving superior source diversity. }
\label{table:GraphSum}
\end{table}

\subsection{QA Accuracy}
 The results of the evaluation of our methodology compared with various RAG baselines can be seen in Table \ref{table:QAAcc}. The largest average improvement is over the Rewrite method and the smallest over RAPTOR. Other than the PubMedQA dataset, our method has comparable or improved performance over the baselines on all datasets. For PubMedQA, we believe that the slight drop in performance is due to insufficient denoising in our created graph, which we plan on addressing in future work. In all, these results imply that our method has allowed the model to more thoroughly analyze the provided data, therefore more effectively synthesizing information from the retrieved documents.

\subsection{Component Level Analyses Results}
We obtained results for each of our relation extraction, graph creation, and graph summarization components. Our relation extraction evaluation compared four methods across three metrics: decontextualization quality (Ref Score), semantic preservation of original documents' meanings (Sem. Similarity), and key claim retention (Claim Ret.), with the results shown in Table \ref{table:Rel}. The entity-based claim extraction approaches (direct\_triples and pairs\_relations) achieved higher reference tracking scores (0.994 and 0.971) compared to claim-based methods (single\_stage 0.941, two\_stage 0.946) due to their focus on extracting explicit entities which naturally avoids leaving unresolved references. However, the claim-based methods still achieved strong semantic preservation performance (0.901 and 0.903 vs 0.865 and 0.815). This advantage suggests that retaining the sentence structure of the claims results in lower information loss of semantic meaning. All of our methods achieved a perfect key claim retention score. These results support our usage of the single stage approach with its comparable decontextualization and superior semantic preservation scores compared to the entity extraction approaches, and it achieves almost identical performance to the two stage approach at a fraction of the computational cost.

For the Graph Creation component, the results of these methods can be seen in Table \ref{table:GraphCrea}. The summaries produced by the graph communities had a higher relevance score to the input question compared to the summaries produced by the semantic communities 59.35\% of the time. While semantic communities are limited to capturing relationships based on pure textual similarity, our graph construction identifies connections that may be relevant topically yet semantically dissimilar. 

For the Graph Summarization component, the results of these metrics can be seen in Table \ref{table:GraphSum}. Our layerwise summarization method achieved comparable faithfulness (0.9569) and relevancy scores (0.8414) compared to the alternative approaches while having superior source diversity (0.9647). The slightly lower relevancy score of our layerwise method (0.8414) compared to semantic clustering (0.8604) stems from the inclusion of information in the summaries that is not directly relevant to the question but is useful for connecting relevant statements. This design decision enables more comprehensive answers but lowers the total number of claims that are directly relevant to the input question in the summaries. The consistently high faithfulness values (>0.94) for all three alternative methods confirms that none of them suffer from significant hallucinations. Our method achieving a strong faithfulness (0.9569) balanced with superior source diversity, means that it can integrate information from many of the retrieved documents with little hallucination in the produced summaries.

\section{Conclusion}

We introduce a novel method for retrieval based BioMedical QA tasks that utilizes propositional claims to construct a local knowledge graph from retrieved documents, before constructing summaries derived via layerwise summarization from the graph to contextualize a small language model to produce the final decisions. We achieved comparable or superior performance with our method over RAG baselines on several biomedical benchmarks, demonstrating its effectiveness. We performed additional experiments covering the intermediate stages of our pipeline, showcasing the robustness of our approach. 

Moving forward, we plan to expand our approach in several directions. First, we plan to enhance our graph creation step's denoising capabilities through dynamic thresholding and improved conflict resolution. Second, we intend to integrate LLM reasoning methods and tool use into our pipeline, potentially improving the QA performance and expanding the number of applicable downstream tasks. Third, we plan to evaluate our approach's generalizability with a wider selection of models, datasets, and benchmarks. Finally, we seek to improve the computational efficiency of our method, focusing on reducing the number of LLM calls for our graph construction and summarization.
\bibliography{custom}

\end{document}